\documentclass[letterpaper, 10 pt, conference]{ieeeconf}  
\IEEEoverridecommandlockouts
\usepackage[ruled,vlined]{algorithm2e}
\usepackage{amssymb}
\usepackage{amsmath, bm} 
\usepackage{mathtools}
\usepackage{multirow}

\title{\LARGE \bf
Differential Dynamic Programming with Nonlinear Safety Constraints Under System Uncertainties
}

\author{Gokhan Alcan and Ville Kyrki
\thanks{This work was financially supported by Academy of Finland (B-REAL project with grant number 328399).}%
\thanks{All authors are with Intelligent Robotics Group, Department of Electrical Engineering and Automation (EEA), Aalto University, Espoo, Finland.
        {\tt\small \{gokhan.alcan, ville.kyrki\}@aalto.fi}}%
}

\usepackage{stackengine}
\newcommand\xrowht[2][0]{\addstackgap[.5\dimexpr#2\relax]{\vphantom{#1}}}
\begin{document}
\setlength{\textfloatsep}{1pt}

\maketitle
\thispagestyle{empty}
\pagestyle{empty}

\begin{abstract}
Safe operation of systems such as robots requires them to plan and execute trajectories subject to safety constraints. When those systems are subject to uncertainties in their dynamics, it is challenging to ensure that the constraints are not violated. In this paper, we propose Safe-CDDP, a safe trajectory optimization and control approach for systems under additive uncertainties and non-linear safety constraints based on constrained differential dynamic programming (DDP). The safety of the robot during its motion is formulated as chance constraints with user-chosen probabilities of constraint satisfaction. The chance constraints are transformed into deterministic ones in DDP formulation by constraint tightening. To avoid over-conservatism during constraint tightening, linear control gains of the feedback policy derived from the constrained DDP are used in the approximation of closed-loop uncertainty propagation in prediction. The proposed algorithm is empirically evaluated on three different robot dynamics with up to 12 degrees of freedom in simulation. The computational feasibility and applicability of the approach are demonstrated with a physical hardware implementation.
\end{abstract}

\section{Introduction}

In many real-world applications, robots are situated in uncertain environments with stochastic dynamics, where they are required to satisfy particular safety constraints such as collision avoidance or physical limits of their actuators. Within this context, \textit{safety} can be defined as the feasibility and stability of a control policy that achieves the requirements of the desired task while satisfying the safety constraints considering uncertainties affecting the system.

Safety in robotics is an active research area and can be studied as trajectory optimization and control under constraints in a stochastic environment. In this context, model predictive control (MPC) is a useful framework, as it allows the optimization of state and input trajectories based on an objective function under chance constraints \cite{heirung2018}. In direct optimization of such a problem, chance constraints are typically transformed to deterministic ones by constraint tightening with precomputed fixed controller gains \cite{hewing2019} or robust constraints are defined using large confidence bounds of uncertainties \cite{ostafew2016}. In both cases, the true effect of feedback on uncertainty propagation is generally omitted, which leads to excessive conservatism and related loss of performance.

\begin{figure}[htbp]
	\centering
	\includegraphics[width=8.8cm]{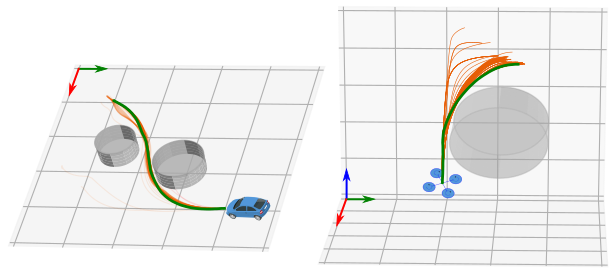}	
	\caption{Safe-CDDP successfully generates optimal trajectories under system uncertainties for complex and underactuated robots. Iteratively generated safe trajectories (orange) and the real trajectories (green) traveled by 2D car-like robot (\textit{Left}) and 3D quadrotor robot (\textit{Right}) in constrained environments.\label{fig-1}}
\end{figure}

To address the question of how to avoid the loss of performance related to conservative constraint tightening, we propose a novel safe trajectory optimization and control approach, which we call Safe Constrained Differential Dynamic Programming (Safe-CDDP). The proposed method extends standard differential dynamic programming (DDP) to handling nonlinear state and input constraints in the presence of additive system uncertainties. The constraints are handled by modeling the problem as a chance-constrained optimal control problem under uncertainty and using constraint tightening \cite{lorenzen2016} to turn it into a deterministic constrained problem, thus generating an appropriate safety margin for the constraints. To reduce the uncertainty in the prediction horizon, we incorporate the control gains of a locally optimal affine feedback policy derived in the backward pass of DDP, which yields a good approximation for optimized trajectories at convergence and avoids excessive conservatism.

To apply the method in real-time feedback control, we also need to address the issue that the tightened constraints are only virtual in the sense that the system noise may cause them to be violated within the safety margin even when they are tightened appropriately. Thus, the utilization of such a constrained formulation in an optimizer creates the risk of a constraint violation, which the optimization process needs to handle appropriately.

The main contributions of our work are:
\begin{enumerate}
	\item A principled way of reflecting the effect of feedback on uncertainty propagation in prediction using dynamic gains from the locally optimal control derived in constrained DDP formulation to avoid excessive conservatism. 
	\item A safe model predictive control approach (Safe-CDDP) with general nonlinear constraints in the presence of additive system uncertainties.	
	\item Simulation studies for three types of robot platform dynamics demonstrating that Safe-CDDP ensures stochastic constraint satisfaction for complex and under-actuated robots in safety-required tasks (Fig. \ref{fig-1}).
	\item Validation via hardware implementation for a differential drive robot that the proposed method is computationally feasible using a low-power CPU.
\end{enumerate}

Compared to existing formulations to similar problems, the proposed approach simultaneously modes the closed-loop effects of additive uncertainties using dynamic gains instead of pre-computed fix gains, and it is not limited to box constraints but includes general nonlinear state and input constraints.

\section{Related Work}

There exist different notions of safety in robotics control literature. In safety-critical control, safety has been investigated through the terms of set-invariance and reachability analysis. A safety set is defined as a set of allowed states for a system and the task of controlling the system becomes ensuring the invariance of this safety set \cite{gurriet2018}.  Reachability of such target sets is formulated as an optimal control problem for safety analysis and synthesis of safe controllers \cite{choi2021}. Even though dynamic programming principles similar to the scope of this paper can be utilized to solve reachability-based value functions, those methods suffer from the usual curse of dimensionality, solving them online on embedded hardware is mostly infeasible and they are generally overly conservative.

Robust model predictive control approaches involve direct optimization that enables to employ of nonlinear state and input constraints by considering the uncertainty with a worst-case safety bound. Since robust MPC is computationally complex, tube MPCs \cite{lopez2019} are designed as an approximation for robust MPCs but they require to define tube geometry dynamics and in some cases time-consuming offline computation \cite{majumdar2017}. Moreover, most of the approaches with fixed tube sizes produce overly conservative tubes due to the lack of knowledge regarding state-dependent uncertainty. Lopez \textit{et al.} \cite{lopez2019} proposed to optimize both the tube geometry and open-loop trajectory simultaneously to reduce conservativeness. However, they assume that the system has the same number of outputs to be controlled as inputs, which makes the method only feasible for relatively low-dimensional and particular kinds of nonlinear systems \cite{lopez2019}. On the other hand, Differential Dynamic Programming (DDP) formulation allows the decomposition of a trajectory optimization problem into smaller ones by limiting the state space to a quadratic trust-region around a current reference solution \cite{lantoine2012theory}, which results in only local optimality but reduces the processing cost dramatically. Although there is no straightforward way to incorporate nonlinear state and input constraints in the DDP formulation, recent attempts are promising. Next, we provide an overview of previous work related to constrained deterministic and stochastic variants of DDPs.

\subsection{Constrained Deterministic DDPs}
\label{CDDP_methods}

Some recent studies target the design of DDP with input constraints \cite{tassa2014}, \cite{sola2020}, \cite{doshi2020}. However, for this work methods that incorporate also state constraints are more relevant. Xie \textit{et al.} \cite{xie2017} utilized Karush-Kuhn-Tucker (KKT) conditions and modified the DDP formulation in the presence of nonlinear state and input constraints according to the active set of constraints at each time step to keep the trajectories feasible. Aoyama \textit{et al.} \cite{aoyama2020} utilized the work developed in \cite{xie2017} and extended it through an Augmented Lagrangian method by considering a set of penalty functions. Pavlov \textit{et al.} \cite{pavlov2021} introduced primal-dual interior-point DDP to handle nonlinear state and input constraints. The method is quite promising since it does not require to identify the active set of constraints, but it needs second-order derivatives of the state constraint. Augmented Lagrangian TRajectory Optimizer (ALTRO) \cite{howell2019} was also recently developed to fuse the advantageous sides of direct methods and DDPs. It handles general deterministic nonlinear state and input constraints with fast convergence and also enables to start with infeasible initial trajectories. 

The aforementioned methods achieve promising performances in handling nonlinear constraints and increase the usability of DDPs in robotics tasks. However, in the presence of system uncertainty, they lack safety provisions, which are the primary issue studied in this work. 

\subsection{Stochastic DDPs}

Todorov and Li \cite{todorov2005} developed input constrained iterative Linear Quadratic Gaussian control for nonlinear stochastic systems involving multiplicative noise with standard deviation proportional to the control signals. Theodorou \textit{et al.} \cite{theodorou2010} then extended it to stochastic DDPs for state and control multiplicative process noise. In their work, the derived control policy does not depend on the statistical characteristics of the noise when the stochastic dynamics have only additive noise. Pan and Theodorou \cite{pan2014} introduced a data-driven probabilistic unconstrained DDP using Gaussian processes to address the uncertainty for dynamics models. Celik \textit{et al.} \cite{celik2019} proposed to model state and action physical limits as probabilistic chance constraints in a DDP structure to avoid catastrophic greedy updates. The method does not apply to tasks that need to employ an arbitrary nonlinear constraints on states such as obstacle avoidance, since it involves only linear box-constraints on states and actions as their limits. Ozaki \textit{et al.} \cite{ozaki2018} proposed a modification for stochastic DDP developed in \cite{theodorou2010} considering the disturbances and the uncertainties as random state perturbations for an unconstrained deterministic dynamical system and employing unscented transform. They then further improved the method by tube stochastic DDP \cite{ozaki2020} to handle the control constraints but nonlinear constraints on states were not included. 

Our approach widens the extent of stochastic DDPs by considering general nonlinear state and input constraints in the presence of additive uncertainties. 

\section{Proposed Method}

We consider a nonlinear discrete-time dynamical system of the form
\begin{equation}
	\label{noisy_transition_dynamics}
	\textbf{\text{x}}_{k+1} = f(\textbf{\text{x}}_k, \textbf{\text{u}}_k) + \bm{\omega}_k,
\end{equation}
where $\textbf{\text{x}}_k\in \mathbb{R}^n$ and $\textbf{\text{u}}_k\in \mathbb{R}^m$ are the system state and the control input at time step $k$, respectively. $f:\mathbb{R}^n\times\mathbb{R}^m$$\rightarrow$$\mathbb{R}^n$ is a nonlinear state transition function which is assumed to be twice differentiable. $\bm{\omega}_k$ is assumed to be spatially uncorrelated independent and identically distributed noise, drawn from a zero mean Gaussian distribution with a known covariance matrix, $\bm{\omega}_k\sim \mathcal{N}(\bm{0},\,\bm{\Sigma}^{\bm{\omega}})$.

For a given initial state $\textbf{\text{x}}_0$, a goal state $\textbf{\text{x}}_{goal}$, and a time horizon $N$, the aim is to find an input trajectory $\{\textbf{\text{u}}_0, ..., \textbf{\text{u}}_{N-1}\}$ that minimizes the expectation of an objective function 
\begin{equation}
	\label{cost_fnc}
	J(\textbf{\text{u}}_0, ..., \textbf{\text{u}}_{N-1}) = \ell^f(\textbf{\text{x}}_N)+\sum_{k=0}^{N-1}\ell(\textbf{\text{x}}_k, \textbf{\text{u}}_k),
\end{equation} 
where $\ell^f:\mathbb{R}^n\rightarrow\mathbb{R}$ and $\ell:\mathbb{R}^n\times\mathbb{R}^m\rightarrow\mathbb{R}$ are the final cost and the running cost, respectively. By considering the input limitations and external constraints on the states, chance-constrained MPC for this problem can be formulated as
\begin{equation}
	\label{chance_cons_MPC}
	\begin{aligned}
		\underset{\textbf{\text{u}}_0, ..., \textbf{\text{u}}_{N-1}}{\text{min}} \quad & \mathop{\mathbb{E}} \Bigg(J(\textbf{\text{u}}_0, ..., \textbf{\text{u}}_{N-1})\Bigg) \\
		\text{subject to} \quad & \textbf{\text{x}}_{k+1} = f(\textbf{\text{x}}_k, \textbf{\text{u}}_k) + \bm{\omega}_k, \\ 
		& \textbf{\text{u}}_{min} \leq \textbf{\text{u}}_k \leq \textbf{\text{u}}_{max}, \\ 
		& \text{Pr}[\bm{g}(\textbf{\text{x}}_{k}, \textbf{\text{u}}_k)\leq \textbf{\text{0}}]>\bm{\beta}_k, \\
		& \textbf{\text{x}}_0 = \bar{\textbf{\text{x}}},
	\end{aligned}
\end{equation}
for all $k = 0, ..., N-1$. Here we simply add box-constraint on control inputs defined by the boundaries $\textbf{\text{u}}_{min}$ and $\textbf{\text{u}}_{max}$. $\bm{g}$ is a vector of $c$ constraints in the form of differentiable functions representing the deterministic state constraints and $\bm{\beta}_k=[\beta_{1,k}, ..., \beta_{c,k}]$ is a vector of minimum satisfaction probabilities for these constraints for time step $k$.

Future predicted states in MPC formulation will result in stochastic distribution due to the noise $\bm{\omega}_k$ in state transition. This leads to chance constraints on states, therefore the safety of the predicted trajectory is determined by the probability of constraint satisfaction. Typically, sufficiently safe control actions are desired to cope with the effect of the uncertainties. To achieve this, chance constraints can be converted into deterministic ones by tightening the constraints \cite{mayne2014}, such that for each original constraint function $g_i(\cdot)$ the tightened constraint is denoted by $\tilde{g_i}(\cdot)$. Overly conservative tightening based on the propagated uncertainty such as employing open-loop uncertainty propagation narrows dramatically the admissible ranges of the states and can make the optimization intractable or even unsolvable \cite{bemporad1999}.

In order to reflect the effect of feedback on uncertainty propagation in prediction and avoid over-conservatism, we propose the following constrained differential dynamic programming-based model predictive control structure. An initial trajectory is optimized in successive backward and forward passes, while the deterministic constraints are updated periodically in constraint tightening. In each backward pass (described in detail in Section III-A), the derivatives of the action-value function ($Q_k(x_k,u_k)$) and optimum value function ($V_k(x_k)$) are calculated backward starting from the final step in the horizon ($N$) using the current estimated (or initialized) trajectories and predefined objective function as in (\ref{cost_fnc}). In forward passes (Section III-B), the nominal state-control trajectory is updated in forward starting from the initial state by using the derivatives obtained in the backward pass and performing local optimizations considering the constraints. In constraint tightening (Section III-C), chance constraints in (\ref{chance_cons_MPC}) are converted into deterministic ones by tightening the constraints using the derived locally optimal control from the backward pass in the uncertainty propagation through prediction. These steps will be detailed in the next sections together with the entire algorithm.

\subsection{Backward Pass}
The backward pass that follows the formulation in \cite{xie2017} with the assumption of deterministic constraints briefly explained here for completeness. Using standard DDP formulation, action-value function $Q_k(\textbf{\text{x}}_k,\textbf{\text{u}}_k)$ can be approximated as a quadratic function as 
\begin{equation}
\label{q_approximation}
\begin{aligned}
Q(\textbf{\text{x}}+\delta\textbf{\text{x}},\textbf{\text{u}}+\delta\textbf{\text{u}}) & \approx Q + Q_\textbf{\text{x}}^{\top}\delta\textbf{\text{x}} +  Q_\textbf{\text{u}}^{\top}\delta\textbf{\text{u}} \\
&+ \frac{1}{2}( \delta\textbf{\text{x}}^{\top}Q_{\textbf{\text{xx}}}\delta\textbf{\text{x}}+ \delta\textbf{\text{u}}^{\top}Q_{\textbf{\text{uu}}}\delta\textbf{\text{u}}) \\ &+ \delta\textbf{\text{x}}^{\top}Q_{\textbf{\text{xu}}}\delta\textbf{\text{u}} 
\end{aligned}
\end{equation}
where $\delta\textbf{\text{x}}$ and $\delta\textbf{\text{u}}$ are the deviations about the nominal action-state pair ($\textbf{\text{x}},\textbf{\text{u}}$). The derivatives of $Q$ function are then
\begin{equation}
\label{q_derivatives}
\begin{aligned}
Q_\textbf{\text{x}} & = \ell_{\textbf{\text{x}}}+f_\textbf{\text{x}}^{\top}V_\textbf{\text{x}}^{\prime}, \\
Q_\textbf{\text{u}} & = \ell_{\textbf{\text{u}}}+f_\textbf{\text{u}}^{\top}V_\textbf{\text{x}}^{\prime}, \\
Q_{\textbf{\text{xx}}} & = \ell_{\textbf{\text{xx}}}+f_\textbf{\text{x}}^{\top}V_{\textbf{\text{xx}}}^{\prime}f_\textbf{\text{x}}+(V_\textbf{\text{x}}^{\prime}f_{\textbf{\text{xx}}}), \\
Q_{\textbf{\text{uu}}} & = \ell_{\textbf{\text{uu}}}+f_\textbf{\text{u}}^{\top}V_{\textbf{\text{xx}}}^{\prime}f_\textbf{\text{u}}+(V_\textbf{\text{x}}^{\prime}f_{\textbf{\text{uu}}}), \\ 
Q_{\textbf{\text{xu}}} & = \ell_{\textbf{\text{xu}}}+f_\textbf{\text{x}}^{\top}V_{\textbf{\text{xx}}}^{\prime}f_\textbf{\text{u}}+(V_\textbf{\text{x}}^{\prime}f_{\textbf{\text{xu}}}).
\end{aligned}
\end{equation}
where $V$ is the value function (see \cite{xie2017, aoyama2020} for details). In order to simplify the notations, we dropped the time step indices, used prime to indicate the next time step and used subscripts for the derivatives. The terms in parentheses in (\ref{q_derivatives}) (second derivatives of the dynamics) provide better local fidelity by capturing the nonlinear effects of the system. Calculation of those tensors can be costly especially for systems with complex dynamics and high-dimensional states such as humanoids, and therefore they can be discarded to obtain faster convergence by sacrificing fidelity, where in that case DDP refers to iterative LQR. The method developed in this study is applicable to both cases.

In unconstrained situations, locally optimal control deviations can be computed by minimizing (\ref{q_approximation}) with respect to $\delta\textbf{\text{u}}$ resulting in
\begin{equation}
\label{optimal_control}
	\delta\textbf{\text{u}}^* = -Q_{\textbf{\text{uu}}}^{-1}(Q_{\textbf{\text{ux}}}\delta\textbf{\text{x}}+Q_\textbf{\text{u}}) \triangleq \bar{\textbf{\text{K}}}\delta\textbf{\text{x}} + \bar{\textbf{\text{d}}}
\end{equation}
where $\bar{\textbf{\text{K}}}$ is the linear feedback gain and $\bar{\textbf{\text{d}}}$ is the affine term. 

Under state and/or input constraints, the optimal control problem turns into
\begin{equation}
\label{cddp_formulation_1}
\begin{aligned}
\underset{\delta\textbf{\text{u}}}{\text{min}} \;\; & Q(\textbf{\text{x}}+\delta\textbf{\text{x}}, \textbf{\text{u}}+\delta\textbf{\text{u}}),  \\ 
\text{subject to} \;\; & \tilde{\bm{g}}(\textbf{\text{x}}+\delta\textbf{\text{x}}, \textbf{\text{u}}+\delta\textbf{\text{u}})\leq 0. 
\end{aligned}
\end{equation}
where the constraint vector $\tilde{\bm{g}}$ includes the deterministic state constraints obtained in the constraint tightening step including input limitations. In order to incorporate the constraints in the estimation of action-value function (\ref{q_approximation}), constraints' first order approximations are obtained as
\begin{equation}
	\begin{aligned}
		\tilde{\bm{g}}(\textbf{\text{x}}+\delta\textbf{\text{x}}, \textbf{\text{u}}+\delta\textbf{\text{u}}) & \approx  \tilde{\bm{g}}(\textbf{\text{x}},\textbf{\text{u}}) \\
		&+\underbrace{\tilde{\bm{g}}_\textbf{\text{u}}(\textbf{\text{x}},\textbf{\text{u}})}_{\triangleq \textbf{\text{C}}}\delta\textbf{\text{u}}+\underbrace{\tilde{\bm{g}}_\textbf{\text{x}}(\textbf{\text{x}},\textbf{\text{u}})}_{\triangleq -\textbf{\text{D}}} \delta\textbf{\text{x}}
	\end{aligned}
\end{equation}

Xie \textit{et al.} \cite{xie2017} proposed to check the set of active constraints during the iterations to make sure that the analytical approximation of the value function around the nominal trajectories still yields a good approximation. To achieve this, all active constraints are included in a set which contains all equality constraints that are always active and inequality constraints for which the value is greater than -$\epsilon$ ($\tilde{g}_i>-\epsilon$) where $\epsilon$ is a small constant for numerical stability.

The optimization problem in (\ref{cddp_formulation_1}) is then converted to
\begin{equation}
\label{cddp_formulation_2}
\begin{aligned}
\underset{\delta\textbf{\text{u}}}{\text{min}} \;\;\; & Q_\textbf{\text{u}}^{\top}\delta\textbf{\text{u}}+\frac{1}{2}(\delta\textbf{\text{u}}^{\top}Q_{\textbf{\text{uu}}}\delta\textbf{\text{u}})+\delta\textbf{\text{x}}^{\top}Q_{\textbf{\text{xu}}}\delta\textbf{\text{u}} ,  \\ 
\text{subject to} \;\;\; & \textbf{\text{C}}\delta\textbf{\text{u}} = \textbf{\text{D}} \delta\textbf{\text{x}}. 
\end{aligned}
\end{equation}
An analytical solution to this problem through KKT conditions \cite{nocedal2006} can be found by solving the pair of equations
\begin{equation}
\label{kkt_prob}
\begin{aligned}
Q_{\textbf{\text{uu}}}\delta\textbf{\text{u}}+Q_{\textbf{\text{ux}}}\delta\textbf{\text{x}}+Q_\textbf{\text{u}}+\textbf{\text{C}}^{\top}\bm{\lambda} & =\textbf{\text{0}}  \\ 
\textbf{\text{C}}\delta\textbf{\text{u}} -\textbf{\text{D}}\delta\textbf{\text{x}} & = \textbf{\text{0}}
\end{aligned}
\end{equation}
where $\bm{\lambda}$ is a vector of Lagrangian multipliers. By solving (\ref{kkt_prob}) with pruned matrices $\tilde{\textbf{\text{C}}}$ and $\tilde{\textbf{\text{D}}}$ that ensures nonexistence of negative $\bm{\lambda}$ values (check \cite{xie2017} for details), locally optimal input deviations can be again expressed as a function of state deviations as follows:
\begin{equation}
\label{cddp_affine_policy}
	\delta\textbf{\text{u}}^* = \textbf{\text{K}}\delta\textbf{\text{x}} + \textbf{\text{d}}
\end{equation}
where the control parameters $\textbf{\text{K}}$ and $\textbf{\text{d}}$ are now calculated as
\begin{equation}
\label{constrained_ctrl_params}
\begin{aligned}
\textbf{\text{K}} = & -Q_{\textbf{\text{uu}}}^{-1}Q_{\textbf{\text{ux}}} \\
& -Q_{\textbf{\text{uu}}}^{-1}\tilde{\textbf{\text{C}}}^{\top} (\tilde{\textbf{\text{C}}}Q_{\textbf{\text{uu}}}^{-1}\tilde{\textbf{\text{C}}}^{\top})^{-1}\tilde{\textbf{\text{D}}} \\
& +Q_{\textbf{\text{uu}}}^{-1}\tilde{\textbf{\text{C}}}^{\top} (\tilde{\textbf{\text{C}}}Q_{\textbf{\text{uu}}}^{-1}\tilde{\textbf{\text{C}}}^{\top})^{-1}\tilde{\textbf{\text{C}}}Q_{\textbf{\text{uu}}}^{-1}Q_{\textbf{\text{ux}}} \;, \\
\textbf{\text{d}} = & - Q_{\textbf{\text{uu}}}^{-1}\Big( Q_\textbf{\text{u}} -\tilde{\textbf{\text{C}}}^{\top}(\tilde{\textbf{\text{C}}}Q_{\textbf{\text{uu}}}^{-1}\tilde{\textbf{\text{C}}}^{\top})^{-1}\tilde{\textbf{\text{C}}}Q_{\textbf{\text{uu}}}^{-1}Q_\textbf{\text{u}} \Big).
\end{aligned}
\end{equation}

It is observed that $\delta\textbf{\text{u}}^*$ at time step $k$ requires the knowledge of first and second derivatives of the value function for next time step ($k+1$) as shown in (\ref{q_derivatives}). By plugging the optimal control found in (\ref{cddp_affine_policy}) into the approximated action-value function (\ref{q_approximation}), the value function can be approximated  as 
\begin{equation}
	V \approx \frac{1}{2}\delta\textbf{\text{x}}^{\top}V_{\textbf{\text{xx}}}\delta\textbf{\text{x}} + V_\textbf{\text{x}}^{\top}\delta\textbf{\text{x}} + \textbf{\text{c}}
\end{equation}
where $\textbf{\text{c}}$ is a constant term and the derivative terms can be calculated as
\begin{equation}
	\label{v_derivatives}
	\begin{aligned}
		V_\textbf{\text{x}} & = Q_\textbf{\text{x}}+\textbf{\text{K}}Q_\textbf{\text{u}}+\textbf{\text{K}}^{\top}Q_{\textbf{\text{uu}}}\textbf{\text{d}}+Q_{\textbf{\text{ux}}}^{\top}\textbf{\text{d}},  \\
		V_{\textbf{\text{xx}}} & = Q_{\textbf{\text{xx}}}+\textbf{\text{K}}^{\top}Q_{\textbf{\text{uu}}}\textbf{\text{K}}+\textbf{\text{K}}^{\top}Q_{\textbf{\text{ux}}}+Q_{\textbf{\text{ux}}}^{\top}\textbf{\text{K}}.
	\end{aligned}
\end{equation}

$V_x$ and $V_{xx}$ at final step in the horizon ($N$) can be calculated as the first and second derivatives of the final cost function ($\ell^f$), respectively. In this way, derivatives of the $Q$ function (\ref{q_derivatives}) for each step in the predicted trajectory can be consecutively calculated backward starting from the final step.  

\begin{figure}[htbp]		
	\centering	
	\includegraphics[width=7cm]{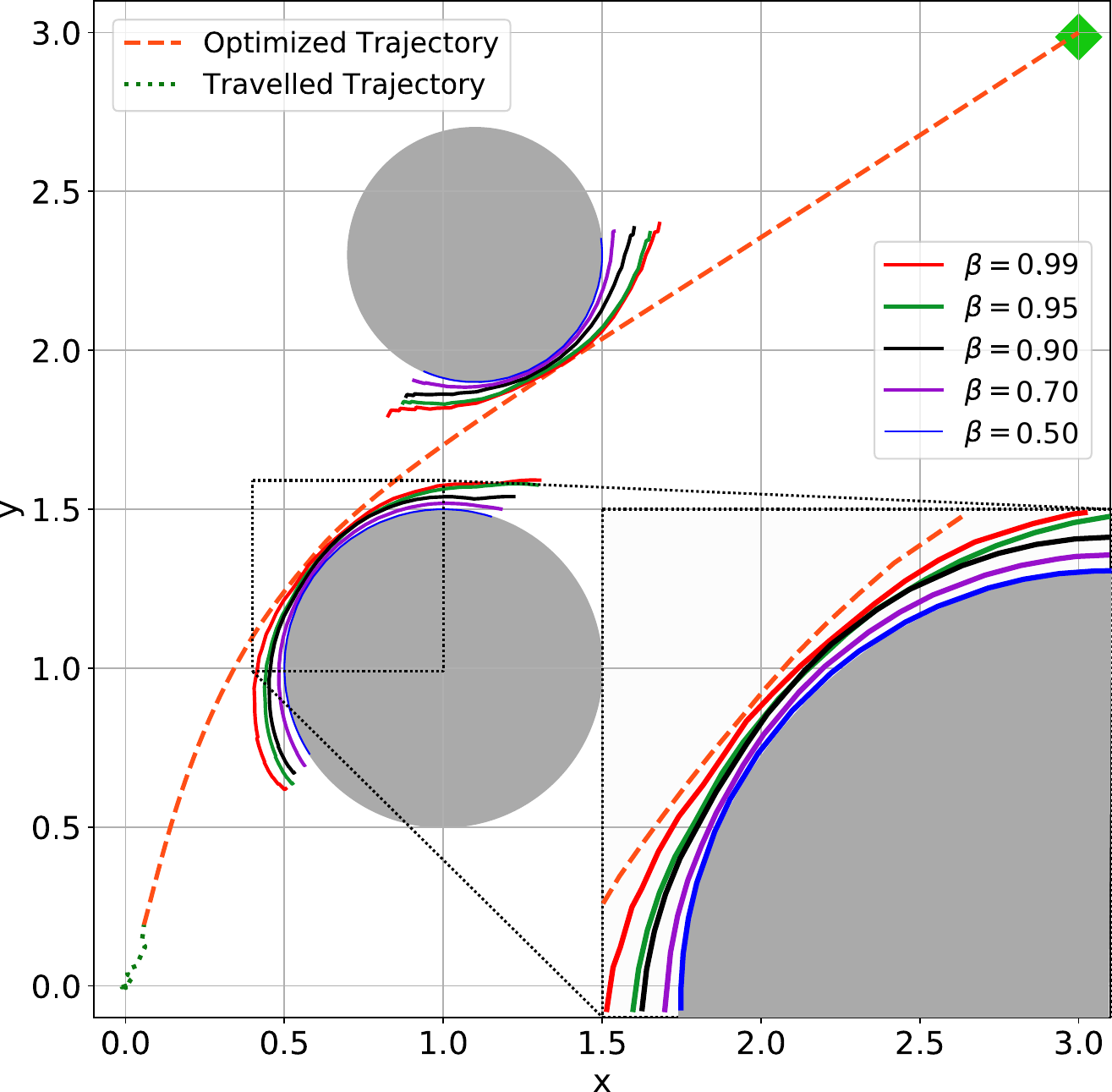}	
	\caption{Changes in the boundaries of the obstacles after tightening the constraints with different $\beta$ values  \label{fig-2}}
\end{figure}

\subsection{Forward Pass}
As the successive step of a backward pass to update the trajectories, forward pass implemented in \cite{xie2017} is briefly explained here for the sake of completeness. 
 
To enforce that the constraints are also satisfied in the forward pass, we formulate the forward pass as the solution of a QP problem
\begin{equation}
	\label{forwardPass}
\begin{aligned}
	\underset{\delta\textbf{\text{u}}}{\text{min}} \;\;\; & Q_\textbf{\text{u}}^{\top}\delta\textbf{\text{u}}+\frac{1}{2}(\delta\textbf{\text{u}}^{\top}Q_{\textbf{\text{uu}}}\delta\textbf{\text{u}})+\delta\textbf{\text{x}}^{\top}Q_{\textbf{\text{xu}}}\delta\textbf{\text{u}} ,  \\ 
	\text{subject to} \;\;\; & \tilde{\bm{g}}(\textbf{\text{x}}+\delta\textbf{\text{x}}, \textbf{\text{u}}+\delta\textbf{\text{u}})\leq 0, \\
	& \textbf{\text{u}}_{min} \leq \textbf{\text{u}}+\delta \textbf{\text{u}}\leq \textbf{\text{u}}_{max} 
\end{aligned}.
\end{equation}
The solution of the problem will then represent the optimal deviation in control input $\delta u$ while respecting the constraints.

The state is then updated by forward integration using the optimal input deviation $\delta \textbf{\text{u}}^*$ as
\begin{equation}
	\textbf{\text{x}}_{k+1} = f(\textbf{\text{x}}_k, \textbf{\text{u}}_k+\delta \textbf{\text{u}}^*)
\end{equation}

\subsection{Constraint Tightening}
\label{constraint_tighten}

This section explains how the chance constraints formulated in (\ref{chance_cons_MPC}) are transformed into the deterministic ones which are required for backward and forward passes. In order to obtain the constraints in a deterministic form, we tighten the constraints using the propagated uncertainty through prediction, which can be decreased efficiently once a feedback policy is employed in the optimization.

Nonlinear transition dynamics $f(\cdot)$ can be approximated to first order by
\begin{equation}
	\textbf{\text{x}}_{k+1} \approx \textbf{\text{x}}_k + f_\textbf{\text{x}}\delta \textbf{\text{x}} + f_\textbf{\text{u}} \delta \textbf{\text{u}}
\end{equation}
Employing the locally optimal control from (\ref{cddp_affine_policy}) results in
\begin{equation}
	\textbf{\text{x}}_{k+1} \approx \textbf{\text{x}}_k + f_\textbf{\text{x}}\delta \textbf{\text{x}} + f_\textbf{\text{u}} \textbf{\text{K}}\delta \textbf{\text{x}}+\textbf{\text{d}} = \textbf{\text{x}}_k + (f_\textbf{\text{x}} + f_\textbf{\text{u}} \textbf{\text{K}})\delta \textbf{\text{x}}+\textbf{\text{d}}
\end{equation}
Closed-loop uncertainty propagation through prediction for the system in (\ref{noisy_transition_dynamics}) then can be approximated as:
\begin{equation}
\label{uncertaintyPropagation}
\bm{\Sigma}^{\textbf{\text{x}}}_{k+1} = (f_\textbf{\text{x}}+f_\textbf{\text{u}}\textbf{\text{K}}_{k})^{\top}\bm{\Sigma}^{\textbf{\text{x}}}_{k}(f_\textbf{\text{x}}+f_\textbf{\text{u}}\textbf{\text{K}}_{k})+\bm{\Sigma}^{\bm{\omega}}_{k+1}
\end{equation}

Since constrained DDP performs a local optimization at each step through prediction, general nonlinear constraint can be locally interpreted as a half-space constraint for that particular step. By utilizing the relationship between probabilistic and robust invariant sets presented in \cite{lorenzen2016} for linear time-invariant systems, the effect of the propagated uncertainty could be approximately taken into account in constraint definitions for each step as follows:
\begin{equation}
\label{const_Tighten}
\tilde{\bm{g}}(\textbf{\text{x}},\textbf{\text{u}}, \bm{\Sigma}^{\textbf{\text{x}}})\triangleq\bm{g}(\textbf{\text{x}},\textbf{\text{u}})+\phi^{-1}(\bm{\beta})\sqrt{\tilde{\bm{g}}_x^{\top}\bm{\Sigma}^{\textbf{\text{x}}}\tilde{\bm{g}}_\textbf{\text{x}}}
\end{equation} 
where $\phi^{-1}$ is the quantile function of the standard normal distribution. Fig. \ref{fig-2} shows that the additional part of the right-hand side in (\ref{const_Tighten}) serves for altering the boundaries of the obstacles and generates safety margins based on the propagated uncertainty and the dynamics of the robot. Moreover, the $\beta$ value adjusts the amount of uncertainty to be considered, which proportionally increases the safety limits.  

Although different $\beta$ values can be assigned for different constraints at different time steps, we did not find it beneficial for safety-required tasks and selected all the $\beta$ values the same for simplicity. 
Once all the chance-constraints are reformulated, the MPC formulation of (\ref{chance_cons_MPC}) can be written in a deterministic form as:
\begin{equation}
	\label{deterministic_MPC}
	\begin{aligned}
		\underset{\textbf{\text{u}}_0, ..., \textbf{\text{u}}_{N-1}}{\text{min}} \quad & \ell^f(\textbf{\text{x}}_N)+\sum_{k=0}^{N-1}\ell(\textbf{\text{x}}_k, \textbf{\text{u}}_k) \\
		\text{subject to} \quad & \textbf{\text{x}}_{k+1} = f(\textbf{\text{x}}_k, \textbf{\text{u}}_k), \\ 
		& \textbf{\text{u}}_{min} \leq \textbf{\text{u}}_k \leq \textbf{\text{u}}_{max}, \\
		& \bm{\Sigma}^{\textbf{\text{x}}}_{k+1} = h(\textbf{\text{x}}_k, \textbf{\text{u}}_k, \textbf{\text{K}}_k, \bm{\Sigma}^{\textbf{\text{x}}}_{k}) \;\; \text{Eq.}(\ref{uncertaintyPropagation}) \\ 
		& \tilde{\bm{g}}(\textbf{\text{x}}_k,\textbf{\text{u}}_k, \bm{\Sigma}^{\textbf{\text{x}}}_k)\leq \textbf{\text{0}}, \\
		& \textbf{\text{x}}_0 = \bar{\textbf{\text{x}}},
	\end{aligned}
\end{equation}
for all $k = 0, ..., N-1$. 

In order to employ a feedback policy in the optimization, Hewing \textit{et al.} \cite{hewing2019} restricted the policy to be a linear state feedback using pre-computed or fixed linear gains. Here, we inherently have a locally optimal affine feedback policy (\ref{cddp_affine_policy}) thanks to the structure of DDP. Moreover, that policy with the controller gains derived in the backward pass (\ref{constrained_ctrl_params}) indeed yields a good approximation for overall nonlinear constrained DDP optimization at convergence. Consequently, the method neither needs to employ some precomputed fixed gains nor needs to solve again a finite LQR problem around the current nominal trajectory.
\subsection{Algorithm in a Nutshell}

\begin{algorithm}[tpbh]
	\SetKwBlock{FnBP}{\textnormal{\textbf{function }\ttfamily Backward\_Pass}}{}
	\SetKwBlock{FnFP}{\textnormal{\textbf{function }\ttfamily Fordward\_Pass}}{}
	\SetKwBlock{FnCT}{\textnormal{\textbf{function }\ttfamily Constraint\_Tightening}}{}
	\DontPrintSemicolon
	\nl \textbf{Init:} $N, \textbf{\text{x}}_0, \textbf{\text{x}}_{goal}, \textbf{\text{x}}_{1,...,N}, \textbf{\text{u}}_{0,...,N-1}$  \;
	\nl \While{robot is not close to $\textbf{\text{x}}_{goal}$}{
		\nl \For{$iter=0, ... ,n_1$}{
			\nl \FnBP{ 
				\nl $V_{x}(\textbf{\text{x}}_N) \leftarrow \ell_{x}^f(\textbf{\text{x}}_N)$\;
				\nl $V_{xx}(\textbf{\text{x}}_N) \leftarrow \ell_{xx}^f(\textbf{\text{x}}_N)$\;
				\nl \For{$k=N-1, ... ,$}{
					\nl Calculate derivatives of $Q$ (Eq. \ref{q_derivatives}) \;
					\nl Calculate derivatives of $V$ (Eq. \ref{v_derivatives}) \; 
					\nl Save $Q_{uu}, Q_{xu}, Q_{u}, K$\;
				}
			}
			\nl \FnFP{ 
				\nl $\textbf{\text{x}} \leftarrow \textbf{\text{x}}_0$ \;
				\nl \For{$k=0, ... , N-1$}{
					\nl $\delta\textbf{\text{x}} \leftarrow \textbf{\text{x}}-\textbf{\text{x}}_k$\;
					\nl Solve QP for $\delta\textbf{\text{u}}$ (Eq. \ref{forwardPass}) \; 
					\nl \eIf{feasible solution}{
						\nl $\textbf{\text{x}}_{k+1} \leftarrow f(\textbf{\text{x}}, \textbf{\text{u}}_k+\delta\textbf{\text{u}})$\;
						\nl $\textbf{\text{x}} \leftarrow \textbf{\text{x}}_{k+1}$\;
					}
					{
						\nl \textit{Regularization} \;
						\nl \textbf{break} \;
					}				
				}
			}
			\nl \If{iter $\; \% \; n_2$==0}{
				\nl \FnCT{
					\nl Calculate closed-loop uncertainty propagation (Eq. \ref{uncertaintyPropagation}) \;
					\nl Update constraints (Eq. \ref{const_Tighten}) \;
				}
			}
		}
		\nl Apply $\textbf{\text{u}}_0$ to robot \;					
		\nl $\textbf{\text{x}}_{0} \leftarrow \text{measurement}$\;
		\nl $\textbf{\text{x}}_{1,...,N-1} \leftarrow \textbf{\text{x}}_{2,...,N}$\;
		\nl $\textbf{\text{u}}_{0,...,N-2} \leftarrow \textbf{\text{u}}_{1,...,N-1}$\;
		\nl $N \leftarrow N-1$\;
	}
	\caption{MPC using Safe-CDDP}
\end{algorithm}

Here we describe the Safe-CDDP algorithm as a whole with some practicalities to employ it in an MPC framework. Even though the initialization of such trajectory optimization algorithms is still an open problem, applying an unconstrained DDP for a relatively straightforward temporary goal state is indeed a convenient practice to initialize our method with suboptimal state-input trajectories. After initialization, our method assumes that there exists a sufficiently large safe region around the real goal state ($\textbf{\text{x}}_{goal}$) and implements a model predictive control procedure with decreasing horizon until the robot gets close to the goal state (Algorithm 1). Depending on the state dimensions, horizon length ($N$) and the number of constraints, iteration number ($n_1$) required to optimize the trajectory should be chosen beforehand. 
Instead of a fixed iteration number for trajectory optimization, one can iterate forward and backward passes until a predetermined convergence is achieved. However, similar to the selection of horizon length for most of the trajectory planning algorithms, the sufficient number of iterations ($n_1$) that saves computational resources is also a hyper-parameter that can be tuned by trial and error. In this study, we choose $n_1=10$ and keep it fixed for simplicity.

Solving (\ref{deterministic_MPC}) in a DDP framework revealed a chicken-egg problem, where the closed-loop uncertainty propagation requires the control gains, and the control gains are calculated in the backward pass of DDP constrained by reformulated chance-constraints using uncertainty propagation. To overcome this dilemma, we initially neglect the noise in transition dynamics at the beginning of backward and forward passes and treat the constraints as deterministic by removing the constraint satisfaction probability. Afterward, the deterministic constraints are updated via constraint tightening.

Note that in this framework, the action-value function is approximated up to second order for trajectory optimization, and the first-order approximation of nonlinear transition dynamics is employed in constraint tightening through uncertainty propagation. Frequent updates in closed-loop uncertainty propagation lead to employing control gains that do not yield a good approximation for action-value function, and therefore it causes oscillations in predicted trajectories and prevents the optimizer from converging. Therefore, we suggest a slower update schedule ($n_2=n_1/2=5$ as a rule of thumb) where iterations get settled for better control gains, then the constraints are updated and the estimated trajectory is optimized for updated constraints again before applying the first input to the robot as depicted in Algorithm 1. The hyper-parameters $n_1$ and $n_2$ can be tuned by trial and error depending on the non-linearity of both system dynamics and constraints. The less complexity allows less iteration and frequent constraint updates.

Finally, it should be noted that the regularization scheme (line 19 in Algorithm 1) and the choice of optimization parameters such as step size and trust region in search space affect the convergence of the algorithm. In our implementation, we used a diminishing trust-region for infeasible solutions and ensured the numerical stability of matrix inversions by adding a regularization term (Please refer to \cite{tassa2012, xie2017,aoyama2020} for regularization details). 

Due to the stochastic setting, the constraints are satisfied with the desired probability $\beta$ as shown in (\ref{chance_cons_MPC}). The quadratic programming solver is able to detect that the solution is infeasible such that all constraints cannot be satisfied, thus, the approach is able to detect infeasibility. However, it is possible that the tightened constraints are not satisfied even when the original constraints are, due to noise in the system. If the aforementioned regularization scheme is not enough to retain feasibility, the tightened constraints can be relaxed as long as the safety requirement of the robotic application allows.

\begin{figure*}[t!]
	\centering
	\includegraphics[width=16cm]{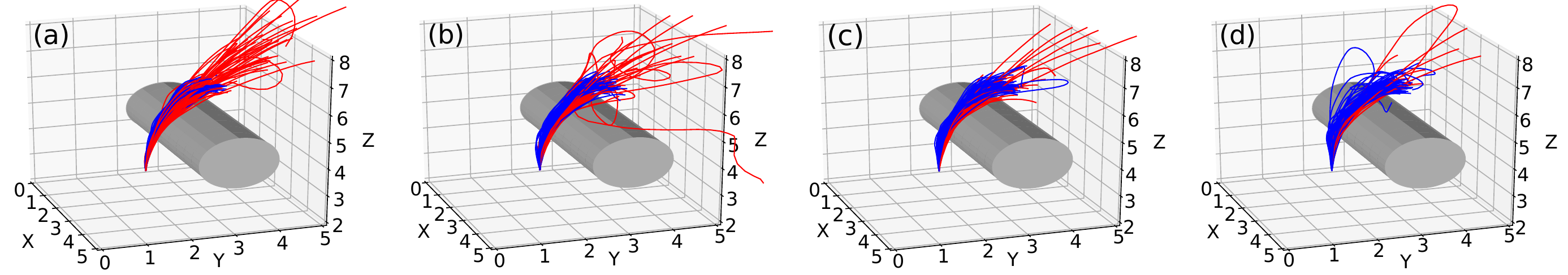}	
	\caption{Comparison of traveled trajectories of a 3D quadrotor in constrainted environment. (a): Classical CDDP without safety precaution. Safe-CDDP with $\beta$ values of (b): 0.90, (c): 0.95 and (d): 0.99. Trajectories in blue represent constraint respected ones and the ones in red for constraint violations.\label{fig-3}}	
\end{figure*}

\section{Experiments}

In this section, we present repetitive simulation results for statistical evaluation and demonstrate hardware implementation to assess the proposed method in the scope of computational feasibility and applicability to real robots.

\subsection{Dynamical Systems in Simulation}

In this section, we statistically evaluate our Safe-CDDP algorithm on three different robot dynamics in simulation: (i) 2D point robot (4 states, 2 inputs)\cite{xie2017}, (ii) 2D car-like robot (4 states, 2 inputs)\cite{xie2017}, and (iii) 3D quadrotor robot (12 states, 4 inputs)\cite{alcan2017, zaki2019}. Prediction horizons ($N$) for the robots are selected as 100, 120 and 50, respectively. The defined task for each robot dynamics is reaching a goal position while avoiding stationary obstacles. The characteristics of process noise ($\bm{\Sigma}_k^{\textbf{\text{x}}}$) is assumed to be known and the positions of the obstacles and the robot itself are assumed to be externally provided without noise. We also assume that there exist safe regions around the initial and goal positions. Once the robot arrived in the safe region around the goal, the task is considered successfully achieved. Another control algorithm could be subsequently triggered to maintain the robot in that region with accurate positioning, which is not a part of this study.

Stationary obstacles in the environment are formulated as circular constraints as defined in \cite{xie2017} and \cite{aoyama2020}. For the 3D quadrotor, it is also possible to use spherical constraints but we preferred to choose cylindrical constraints to test the performance in a more challenging scenario. In addition to obstacle avoidance, i.e., nonlinear state constraints, the limits on control inputs are also employed. Only for the quadrotor robot, Safe-CDDP is implemented in iLQR form for simplicity.

Once the task configuration is fixed for a robot, it was performed repeatedly 100 times for each of 4 different $\beta$ values \{0.5, 0.90, 0.95, 0.99\}. Note that the case with $\beta=0.5$ corresponds to classical CDDP without safety precautions, since $\phi^{-1}(0.5)=0$.

\subsection{Simulation Results}

In order to assess the safety of the obtained trajectories, we utilized 3 metrics regarding constraint violations (Table \ref{violationTable}). Once the robot finishes the task with either success or failure, we call it one episode and repeat the same task with a fixed configuration for 100 episodes. ``\textit{Violated episode}'' refers to an episode in which the robot violates a constraint, \textit{i.e.}, collides with an obstacle, at least once. Average in violated episodes is the ratio of the total number of collisions and the number of violated episodes. Similarly, the total average of violations is the ratio of the total number of collisions and the total number of episodes, \textit{i.e.}, 100 in our case.

\begin{table}[htbp]	
	\caption{Constraint Violation Results \label{violationTable}}
	\begin{center}
		\begin{tabular}{r|c|c|c|c} \hline	\xrowht{4pt}
			&\textbf{CDDP}&\multicolumn{3}{c}{\multirow{2}{*}{\textbf{Safe-CDDP}}} \\
			&\cite{xie2017}\\ \hline 
			\hline	\xrowht{4pt}	
			\textit{\textbf{2D Point Robot}} &  & \textbf{$\beta$=0.90} & \textbf{$\beta$=0.95} & \textbf{$\beta$=0.99} \\ \hline \xrowht{4pt}
			Num. of violated episodes & 64 & 14 & 8 & 0 \\ \hline \xrowht{4pt}
			Avg. in violated episodes & 1.98 & 1.43 & 1.38 & 0 \\ \hline \xrowht{4pt}
			Total avg. of violations & 1.27 & 0.20 & 0.11 & 0 \\ \hline   
			\hline  \xrowht{4pt}  
			\textit{\textbf{Car-like Robot}}&  & \textbf{$\beta$=0.90} & \textbf{$\beta$=0.95} & \textbf{$\beta$=0.99} \\ \hline \xrowht{4pt}  
			Num. of violated episodes & 38 & 2 & 0 & 0 \\ \hline \xrowht{4pt}  
			Avg. in violated episodes & 1.63 & 1.0 & 0 & 0 \\ \hline \xrowht{4pt}
			Total avg. of violations & 0.62 & 0.02 & 0 & 0 \\ \hline   
			\hline  \xrowht{4pt}
			\textit{\textbf{3D Quadrotor Robot}}&  & \textbf{$\beta$=0.90} & \textbf{$\beta$=0.95} & \textbf{$\beta$=0.99} \\ \hline \xrowht{4pt}
			Num. of violated episodes & 66 & 34 & 27 & 15 \\ \hline \xrowht{4pt}
			Avg. in violated episodes & 5.7 & 4.26 & 5.19 & 4.07 \\ \hline \xrowht{4pt}
			Total avg. of violations & 3.76 & 1.45 & 1.4 & 0.61 \\ \hline  
		\end{tabular} 
	\end{center}	
\end{table} 

\begin{figure}[htbp]
	\centering
	\includegraphics[width=8cm]{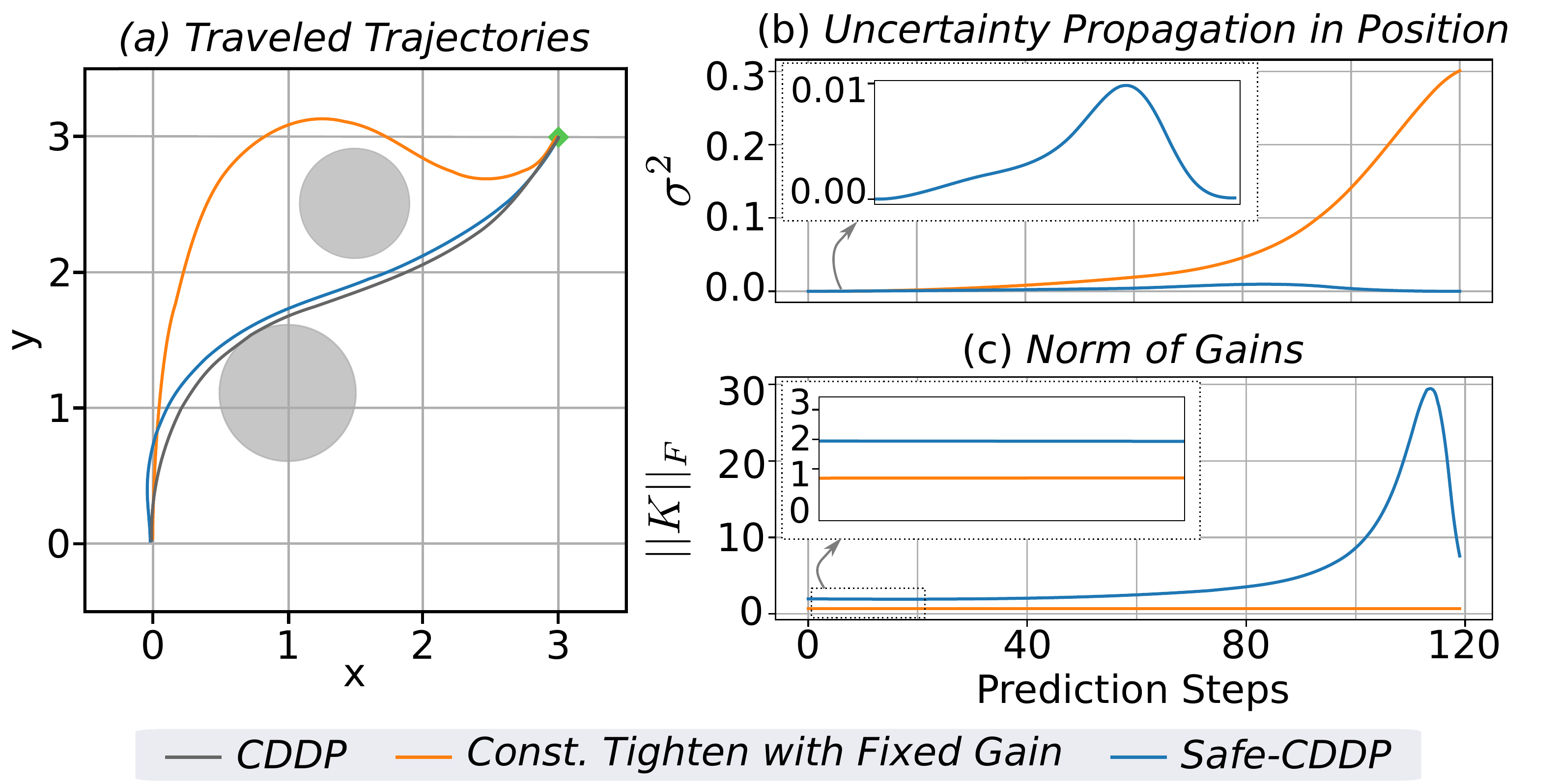}
	\vspace{-10pt}
	\caption{Comparison of CDDP, constraint tightening with fixed gains, and Safe-CDDP for trajectory optimization of a car-like robot in terms of (a) traveled trajectories (b) uncertainty propagation in position, and \linebreak (c) Frobenius norm of gains. \label{fig-4}}
\end{figure}

Table \ref{violationTable} depicts that Safe-CDDP algorithm provides safety in terms of obstacle collision due to uncertainties. By selecting $\beta$=$0.99$, Safe-CDDP performed well for point robot and car-like robot without any constraint violation, where at least one constraint violation occurred in 64 and 38 percent of the tasks for these robots when the safety precaution was removed ($\beta$=0.5). Selecting $\beta$=$0.95$ was even sufficient for the car-like robot to obtain safe trajectories with zero collision. This implies that increasing $\beta$ values, \textit{i.e.}, increasing the confidence bound on the approximated uncertainty propagation leads optimizer to be more conservative and force it to find trajectories away from the obstacles if admissible.

It should be noted that the performance of Safe-CDDP highly depends on the admissible region of states possible for the task definition. For the 3D quadrotor robot, we deliberately select a goal position above the cylinder to assess the capabilities of the method. Fig. \ref{fig-3} presents all the traveled trajectories by quadrotor using Safe-CDDP with different $\beta$ values. Once $\beta$ is set to 0.5, \textit{i.e.}, in the case of classical CDDP, the majority ($\%$66) of the tasks were finished with violations. In Safe-CDDP case with $\beta$=0.99, collision avoidance was not satisfied completely as in the cases of point robot and car-like robot, but it is dramatically improved by diminishing from $\%$66 to $\%$15. This indicates that Safe-CDDP can achieve remarkable performances in decreasing the number of constraint violations for complex and under actuated systems in the presence of system uncertainties. 

\begin{figure}[htbp]
	\centering
	\includegraphics[width=7.8cm]{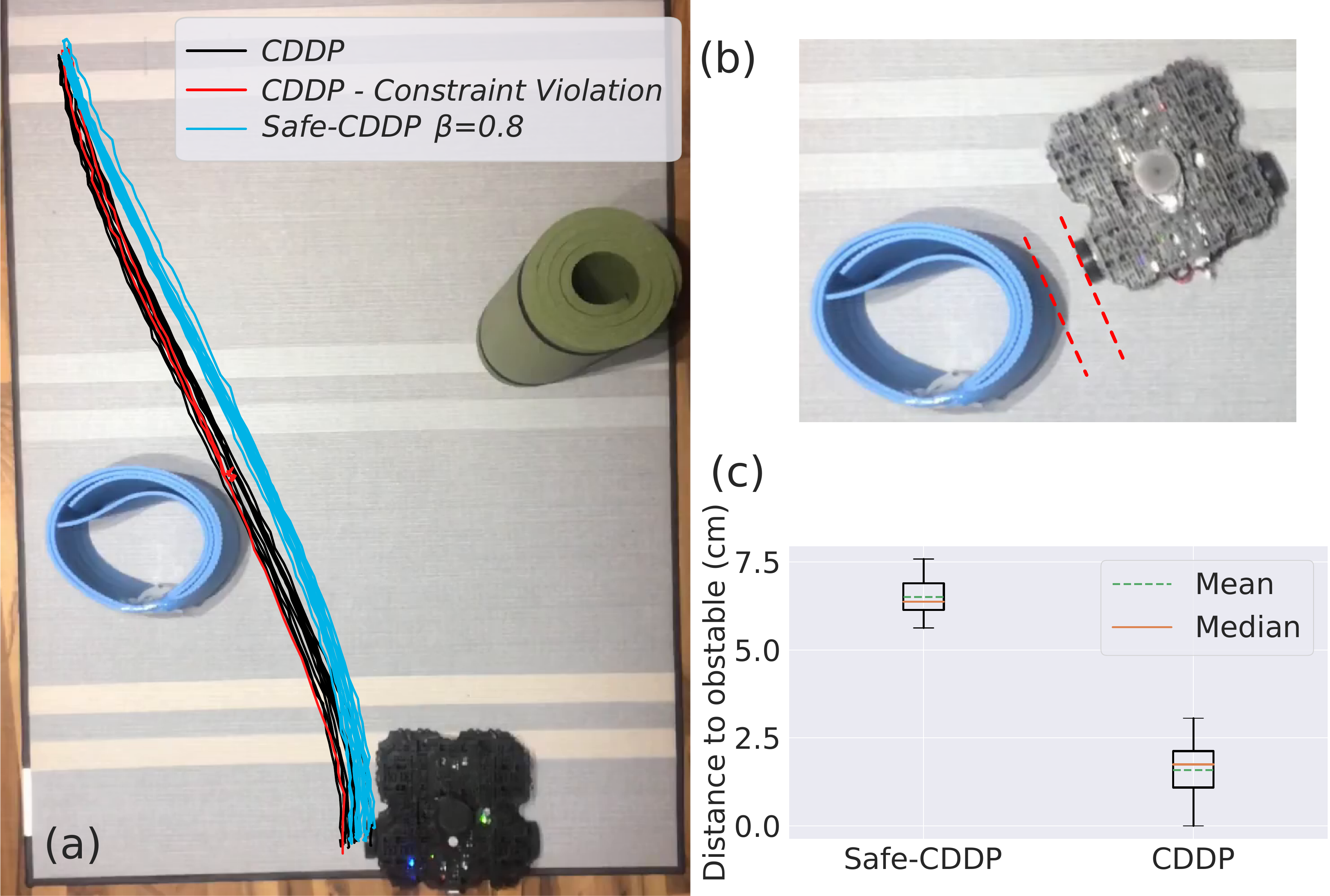}	
	\caption{(a) Comparison of traveled trajectories optimized by CDDP vs Safe-CDDP (b) Safety is determined by the distance between the wheel of Turtlebot and closest obstacle (c) Distributions of distance to closest obstacle\label{fig-5}}
\end{figure}

We further compared our method with the one that uses fixed gains in constraint tightening, similar to the closest collocation method, called Cautious MPC \cite{hewing2019}. An ancillary linear controller was employed in constraint tightening with $\beta=0.99$ for all prediction steps. Even though the fixed controller gains reduced the uncertainty propagation in position up to 40 prediction steps, it started to increase exponentially afterward (Fig. \ref{fig-4}-b). Constraint tightening based on this propagated uncertainty resulted in a more conservative trajectory (Fig. \ref{fig-4}-a). On the other hand, Safe-CDDP avoids excessive conservatism with the same confidence bounds ($\beta=0.99$) by increasing the norm of gains through the prediction horizon appropriately (Fig. \ref{fig-4}-c).

\subsection{Hardware Implementation}
As an experimental setup, we employed Turtlebot3 Waffle Pi that is a differential drive type mobile robot. In order to test Safe-CDDP on Turtlebot, we placed two circular objects with different sizes in the robot's environment as stationary obstacles. Gmapping was used to map the environment and the fusion of Adaptive Monte Carlo Localization (AMCL) and odometry were employed to retrieve the states of the robot (x,y,$\theta$) with an estimation uncertainty. Circular obstacles with the radius of 0.15 and 0.11 m were detected at (x:0.85, y:0) and (x:0.5, y:0.85) locations, respectively. Turtlebot was initially located in ($x$:0, $y$:0, $\theta$:0) and reaching to goal state ($x$:1.4, $y$:0.6, $\theta$:0) via Safe-CDDP was performed with the prediction horizon $N=90$. Standard deviations of position noise retrieved by AMCL were approximately 0.001. 

Safe-CDDP algorithm was implemented as a ROS node in C++ using OSQP \cite{stellato2020} and Eigen \cite{guennebaud2010} libraries. The execution of Safe-CDDP node was tested on both Turtlebot's CPU (Raspberry Pi-4 Model-B with 8GB RAM) and a workstation laptop (Intel i7-6820HQ CPU, 2.70GHz, 8 cores, 16GB RAM). By setting $\beta=0.8$, Turtlebot was able to plan and successfully travel a safe trajectory to the goal position avoiding the obstacles with reasonable safety margins, whereas CDDP results in the trajectories that pass very close to obstacles and sometimes collide with one of the obstacles (Fig. \ref{fig-5}). 

Our proposed approach does not require any additional QPs to be solved compared to CDDP. Only the added computation is in the form of uncertainty propagation and constraint tightening, which corresponds to less than $\%$2 of CDDP execution. The worst-case execution times of a single iteration including backward and forward pass with the trajectory length $N=90$ are approximately 63ms and 22ms for Raspberry Pi and workstation, respectively. Then the execution times decrease dramatically during traveling due to a decrease in active constraints and prediction horizon. This ensures the applicability of the proposed method in real robots with a control frequency faster than 10hz. 

\section{CONCLUSIONS}
We presented a novel safe trajectory optimization approach for nonlinear systems with nonlinear state and input constraints under additive system uncertainties. 
The proposed method reduces over conservatism in the approximation of uncertainty propagation through prediction.
It also enables to impose safety precautions on general nonlinear constraints. 
Computational feasibility and applicability of the proposed approach were shown in hardware implementation, and the effectiveness of the method was validated on three different robot dynamics in simulation. 
 
The proposed approach approximates the future prediction uncertainty by propagating it considering the implicit gain of the DDP feedback controller. 
It is likely that potential approximation errors in this prediction can be compensated conservatively by increasing the safety factor. 
However, an exact solution would be useful to limit the need for this, but
whether it is possible to find such an exact solution remains an interesting open issue. 

Recently, model-based reinforcement learning (RL) has received lots of attention, and integration of optimal control with safety guarantees to systems with learned dynamics seems to offer great possibilities. The proposed approach would be applicable to data-driven models of system dynamics models when their prediction uncertainty is modeled as additive Gaussian noise, which is the case for example for Gaussian Process models. Thus, we believe that the proposed method can be integrated as a part of a RL system that would be able to provide safety guarantees also for its exploration. 

\addtolength{\textheight}{-12cm}   


\end{document}